\documentclass{article}
\usepackage{arxiv}

\usepackage{geometry}
\usepackage{graphicx}
\usepackage[hidelinks]{hyperref}

\usepackage{amssymb}
\usepackage{bm}
\usepackage{lineno}
\usepackage{amsmath}

\usepackage{multirow}
\usepackage{float}
\usepackage{tabularx}
\usepackage{subcaption}
\usepackage{booktabs}
\usepackage{array,xcolor}

\usepackage{algorithm, algpseudocode}

\usepackage{cite}
\title{Neuroscience inspired scientific machine learning (Part-1): Variable spiking neuron for regression}
\author{
  Shailesh Garg  \\
  Department of Applied Mechanics\\
  Indian Institute of Technology Delhi\\
  Hauz Khas, New Delhi 110016, India. \\
  \texttt{shaileshgarg96@gmail.com} \\
  \And
  Souvik Chakraborty  \\
  Department of Applied Mechanics\\
  Yardi School of Artificial Intelligence (YScAI)\\
  Indian Institute of Technology Delhi\\
  Hauz Khas, New Delhi 110016, India. \\
  \texttt{souvik@am.iitd.ac.in}}
\begin{document}
\maketitle
\begin{abstract}
Redundant information transfer in a neural network can increase the complexity of the deep learning model, thus increasing its power consumption. We introduce in this paper a novel spiking neuron, termed Variable Spiking Neuron (VSN), which can reduce the redundant firing using lessons from biological neuron inspired Leaky Integrate and Fire Spiking Neurons (LIF-SN). The proposed VSN blends LIF-SN and artificial neurons. It garners the advantage of intermittent firing from the LIF-SN and utilizes the advantage of continuous activation from the artificial neuron. This property of the proposed VSN makes it suitable for regression tasks, which is a weak point for the vanilla spiking neurons, all while keeping the energy budget low. The proposed VSN is tested against both classification and regression tasks. The results produced advocate favorably towards the efficacy of the proposed spiking neuron, particularly for regression tasks.
\end{abstract}
\section{Introduction}
The introduction of Artificial Neurons (ANs) \cite{mcculloch1943logical} sparked a revolution in the research of Artificial Intelligence (AI), and its adoption increased meteorically with the introduction of the backpropagation algorithm \cite{rumelhart1986learning,jain1996artificial}. While the AN was inspired by how the brain processes information, its mechanics are very rudimentary, and as such, its energy budget is colossal compared to that of the brain. Brain \textit{in vivo} has shown tremendous computational abilities while maintaining a meager energy budget \cite{attwell2001energy,balasubramanian2021brain}. The brain comprises of vast interwoven nets of biological neurons which transfer huge amounts of data between them. Despite having billions of interconnected neurons, the brain can work with such a meager energy budget because the information between these neurons is transferred intermittently, as and when required, in an all or none fashion \cite{barnett2007action}. Thus, only a finite number of neurons are engaged while performing a specific task. The information transfer in ANs, on the other hand, is continuous, and hence, their energy consumption is massive.

One potential alternative to ANs is the Spiking Neurons (SNs) \cite{izhikevich2003simple,pfeiffer2018deep}, where the neurons mimic the behavior of the biological neuron models. SNs collect the incoming information and pass that information only when a predetermined threshold is crossed. The networks utilizing SNs are termed Spiking Neural Networks (SNNs) \cite{ghosh2009spiking}. Because of integrate-and-fire behavior, SNs are energy efficient and consume less energy. The mathematical models for biological neurons available in the literature often have an inverse relationship between simplicity and biological plausibility \cite{yamazaki2022spiking,izhikevich2004model}. The Hodgkin-Huxley model \cite{hodgkin1952quantitative} is among the most biologically plausible, but it is too intricate to implement for deep learning \cite{lecun2015deep} architectures, whereas the artificial neuron model, which is among the least biologically plausible models, has proven to be instrumental in the field of deep learning. In SNNs, the Leaky Integrate-and-Fire (LIF) \cite{burkitt2006review,yamazaki2022spiking,long2010review} model is among the most popular SN models due to the excellent balance between its simplicity and biological plausibility. 

In recent times, SNNs have seen massive growth as the technology paradigm shifts towards a more energy-efficient future. Their inherent energy-saving properties, coupled with technologies like neuromorphic hardware \cite{schuman2017survey,davies2021advancing}, have brought down the resource requirements of deep learning algorithms significantly. SNNs thus have the potential to better integrate AI with physical technology, especially for applications like robotics \cite{raj2019primer}, drones \cite{hernandez2021ai,rawat2023ai}, medical devices \cite{hamet2017artificial}, etc., where there is often a power constraint associated. With all its benefits and potential, it is to be noted that a large part of SNN's growth comes from fields of research heavily dependent on classification tasks. The accuracy of SNNs in classification tasks \cite{kim2020spiking,yan2021energy,dora2016development,ghosh2009new,ponulak2010supervised} is now comparable to that of Artificial Neural Networks (ANNs) for both ANN-to-SNN converted \cite{deng2021optimal,liu2022spikeconverter} networks or natively trained SNNs \cite{neftci2019surrogate,eshraghian2021training}. This is also expected, as the SNNs are trying to emulate the human brain, and as far as the brain is concerned, it is also excellent in classification tasks. The story breaks, however, when it comes to prediction or regression tasks. To the best of the authors' knowledge, the research in the field of SNNs for regression tasks \cite{henkes2022spiking,gehrig2020event,kahana2022function,zhang2022sms,gruning2014spiking}, is at its nascent stages, despite SNNs being decades-old technology.

In this paper, we present a novel spiking neuron that has the advantage of integrate-and-fire similar to a LIF neuron in SNNs and has the advantage of linear/nonlinear continuous activation similar to that observed in AN. The developed novel spiking neuron is termed Variable Spiking Neuron (VSN), and the deep learning architectures using VSN are termed Variable Spiking Neural Networks (VSNNs). The key properties of the proposed VSNs can be summarised as follows:
\begin{enumerate}
    \item \textbf{Intermittent communication:} Contrary to prevailing ANs, the communication in proposed VSN is sparse because of its \textit{integrate and fire} routine while processing the information. 
    \item \textbf{Regression capabilities:} VSNNs, utilizing the proposed VSN, have shown tremendous performance in regression tasks that far outweigh the performance of LIF-SNs and is comparable to that of ANs. This is made possible because of its ability to incorporate \textit{continuous activations} while promoting sparsity.
    \item \textbf{Simplicity while maintaining plausibility:} The proposed VSN displays excellent performance and biological plausibility while retaining implementation simplicity, which is key for both energy saving and flexible implementation in complex deep learning architectures. 
    \item \textbf{Energy saving:} Because of its sparse communication, the VSNNs utilizing the proposed VSN are more energy efficient than the ANNs in both classification and regression tasks. Furthermore, this energy efficiency comes at negligible loss of accuracy, if any.  
\end{enumerate}

To test the efficacy of VSN, we have shown two classification examples and two nonlinear regression examples, tackling various datasets with single and multiple input features. The results produced indicate that the proposed VSN is at par with ANs and SNs in classification tasks and is superior to SNs for regression tasks.

The rest of the paper is arranged as follows, section \ref{section: background} discusses the background and details the mechanisms for biological neurons, ANs, and SNs. Section \ref{section: Prob Statement} discusses the problem statement, and section \ref{section: VSNN} details the proposed VSN. Section \ref{section: case studies} discusses the various examples, and section \ref{section: conclusion} concludes the findings along with the future scope of this research.
\section{Background}\label{section: background}
\begin{figure}[ht!]
    \centering  
    \includegraphics[width=0.75\linewidth]{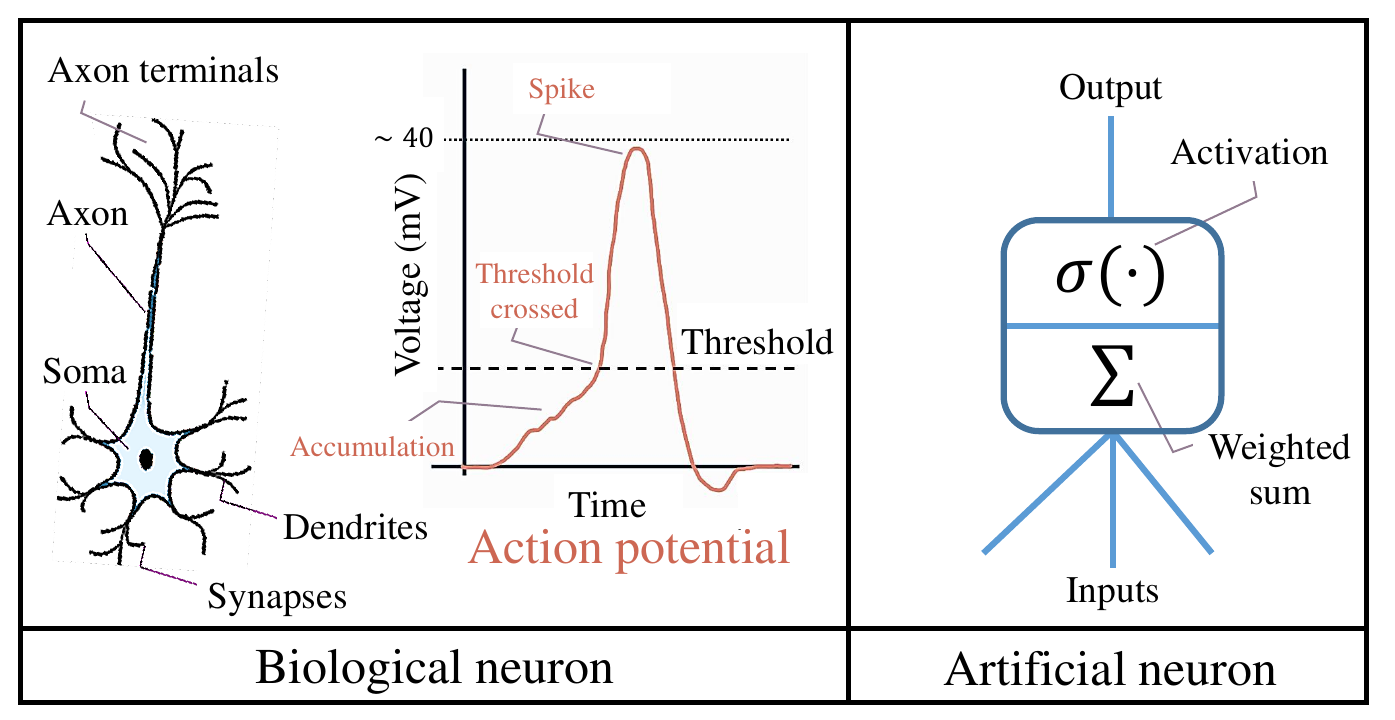}
    \caption{Schematic for biological neuron and artificial neuron.}
    \label{figure: fc 1}
\end{figure}

\noindent \textbf{Biological mechanism for learning:} Information in our body is transferred through a vast network of biological neurons, termed the nervous system. Similarly, the brain also has an interwoven net of biological neurons, which aid the brain in learning and performing tasks. When a task is being performed inside the brain, among billions of biological neurons, only a portion of them are activated to perform the task. The exact quantity of neurons being activated will depend on the task at hand. This allows the brain to conserve energy and perform huge tasks with a meager energy budget. This selective activation is largely possible due to the biological neuron's capacity for integrating the incoming information and sending information forward only when the integrated information crosses a threshold. Once the neuron transfers the information stored, it resets its voltage, and the whole process is started again. The process of accumulation of information, firing, and resetting is collectively known as an action potential \cite{barnett2007action}. A schematic for the biological neuron and the action potential is shown in Fig. \ref{figure: fc 1}.\\

\noindent\textbf{Artificial neuron:} AN's basic premise is that it accepts incoming information from various sources, takes its weighted sum, combines it with its own bias, and passes the whole sum through an activation function. Mathematically, this can be written as,
\begin{equation}
\begin{aligned}    
    z &= \sum\limits_iw_ix_i+b,\\
    y &= \sigma(z),
\end{aligned}
\end{equation}
where $x_i$ are the incoming inputs, $w_i$ are the weights assigned to each input, $b$ is the AN's bias and $\sigma(\cdot)$ is its activation function. The activation function can be linear or nonlinear, continuous or discontinuous. However, practically, ANs utilize only continuous activation to leverage the backpropagation algorithm for learning. The AN learns any task by changing its weights and biases. A schematic for the AN is shown in Fig. \ref{figure: fc 1}.

\noindent\textbf{Spiking neuron:} SNs emulate the biological neurons and use intricate neuron models to achieve the same. Several mathematical models \cite{long2010review,mishra2006exploring} exist for biological neurons that model the action potential using various differential/algebraic equations. Among these, the Hodgkin-Huxley model \cite{hodgkin1952quantitative} is widely accepted as the most biologically plausible, but its comprehensive detailing makes it impractical for use in deep learning architectures. Neuron models like the Izhikevich model \cite{izhikevich2003simple}, the Wilson-Cowan model \cite{wilson2021evolution}, and the Fitzhugh Nagumo model \cite{izhikevich2006fitzhugh,long2010review} are also viable options, but the most widely adopted SN model in literature is the LIF model \cite{burkitt2006review,yamazaki2022spiking,long2010review}. It is sufficiently biologically plausible and, at the same time, trivial enough to implement in deep learning architectures. Mathematically the digital implementation of the LIF neuron model can be described as follows,
\begin{equation}
\begin{aligned}
    M^{(\tilde t)} &= \beta M^{(\tilde t-1)} + z^{(\tilde t)},\\
    y^{(\tilde t)} &= \left\{\begin{array}{ll}
    1;& \,\,\,\,\, M^{(\tilde t)}\ge T\\
    0; & \,\,\,\,\, M^{(\tilde t)}<T\end{array}\right.,
    \text{\,\,\,\,\,if } y^{(\tilde t)} = 1, M^{(\tilde t)} \leftarrow 0
\end{aligned}
\label{equation: LIF eqns}
\end{equation}
where $z^{(\tilde t)}$ is the input from the previous layer, $\beta$ is the leakage parameter, $T$ is the threshold of the neuron, and $M^{(\tilde i)}$ is the memory of neuron for the $i^{\text{th}}$ Spike Time Step (STS). The equation shown is for computing the output spike of $t^{\text{th}}$ STS. The \textit{STS} here refers to the time step of the spike train entering into the neuron and exiting out of the neuron. In Eq. \eqref{equation: LIF eqns}, $\beta$ and $T$ can also be treated as trainable parameters in addition to other trainable parameters of the NN. Now, because there is a discontinuity involved in the computation of the output spike train, we can no longer use the backpropagation algorithm in its vanilla form. Hence, to natively train SNNs, surrogate backpropagation \cite{neftci2019surrogate,eshraghian2021training} can be used. A schematic for the information flow in SNs, specifically LIF neurons, is shown in Fig. \ref{figure: fc 2}.\\

\section{Problem statement}\label{section: Prob Statement}
The artificial neurons and spiking neurons discussed in the previous sections each present some advantages and challenges. The key challenges being tackled here are as follows:
\begin{enumerate}
    \item \textbf{Energy consumption of artificial neurons:} ANs, while excellent in both classification and regression tasks, consume vast amounts of energy, thus limiting their applications to tasks where there are no limits on available energy.
    \item \textbf{Regression performance of spiking neurons:} SNs can deal with the energy challenge discussed above in their vanilla form, but their application is limited to classification tasks. Their application in regression tasks, which are often encountered in engineering domains, is yet to be fully explored owing to their poor performance for such tasks.
\end{enumerate}
Any proposed neuron model should be able to tackle the issues discussed above, i.e., it should be more energy efficient than the artificial neuron and should cater to both classification and regression tasks. Apart from these challenges, a minor yet important point of concern is that the developed spiking neuron model should be able to perform well with direct inputs. This is important because, in regression tasks, we require high-precision encoding, which requires the use of multiple STS. This can be detrimental from an energy consumption point of view, and hence, it is preferred to use direct inputs and allow the network to encode them as per the requirement.
\section{Variable Spiking Neurons}\label{section: VSNN}
\begin{figure*}[ht!]
    \centering    \includegraphics[width=0.9\linewidth]{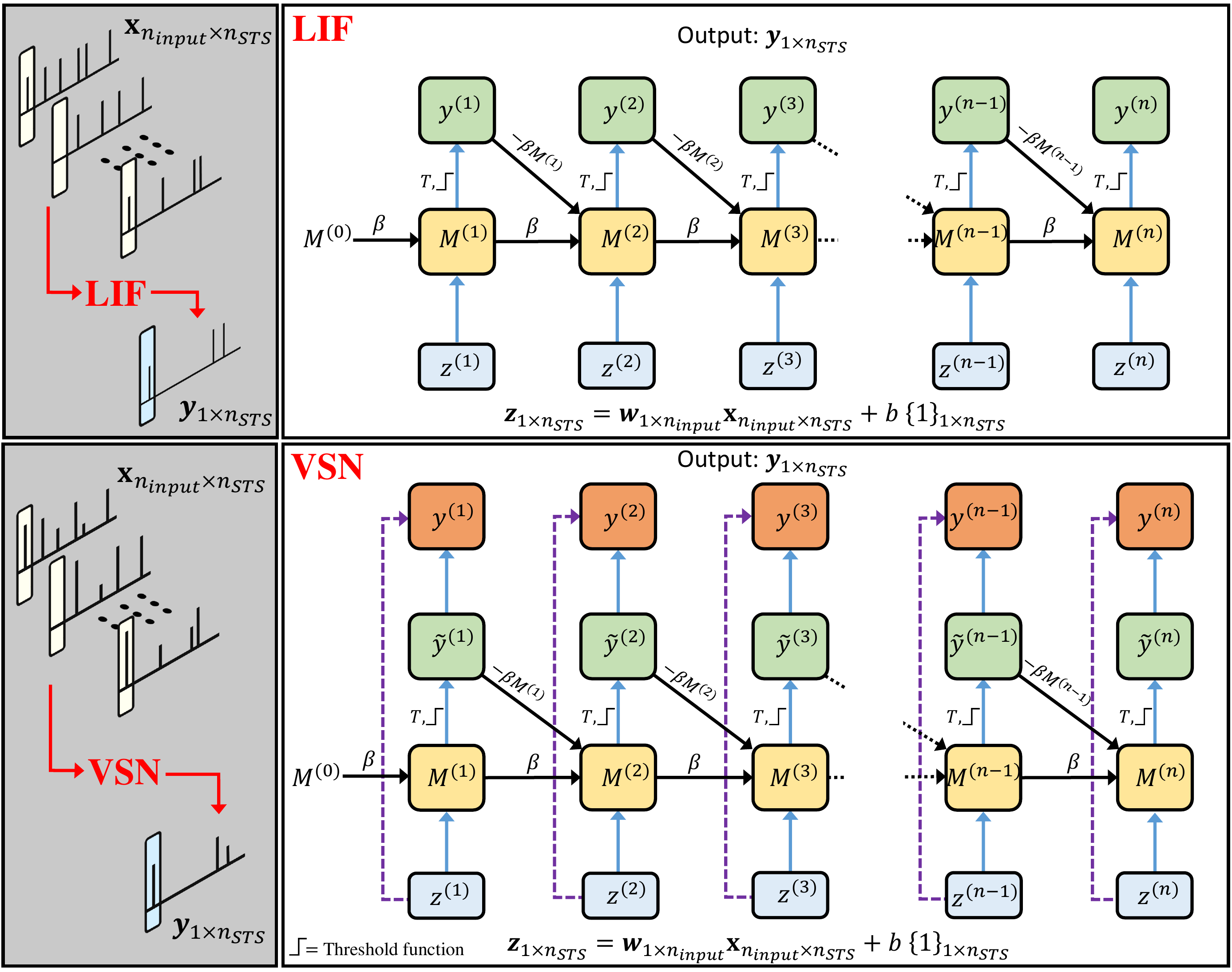}
    \caption{Information flow in forward pass of the spiking neurons.}
    \label{figure: fc 2}
\end{figure*}

In this paper, we present a novel variable spiking neuron model, which is an amalgamation of LIF-SN and AN. The existing ANNs, which work excellently for regression and classification tasks, also have a large number of neurons, but not all pass along useful information, which was the biggest cause of excessive energy consumption in ANs. Therefore, the idea behind VSN is to enable the storage capacity in an AN similar to that observed in biological neurons. To achieve this, the inspiration is taken from the LIF neuron. The mathematics behind the proposed VSN is described as follows,
\begin{equation}
\begin{aligned}
    M^{(\tilde t)} &= \beta M^{(\tilde t-1)} + z^{(\tilde t)},\\
    \widetilde y^{(\tilde t)} &= \left\{\begin{array}{ll}
    1;&\,\,\,\,\, M^{(\tilde t)}\ge T\\
    0;&\,\,\,\,\, M^{(\tilde t)}<T\end{array}\right.,\text{\,\,\,\,\,if } \tilde y^{(\tilde t)} = 1, M^{(\tilde t)}  \leftarrow 0\\
    y^{(\tilde t)} &= \sigma(z^{(\tilde t)}\tilde y^{(\tilde t)}), \,\,\,\,\,\text{given, }\sigma(0) = 0, 
\end{aligned}
\label{equation: VSN eqns}
\end{equation}
where $\sigma(\cdot)$ is a continuous activation such that $\sigma(0)=0$. The term $\sigma(z^{\tilde t}\tilde y^{\tilde t})$ coupled with the constraint placed on $\sigma(\cdot)$ shows that the continuous activation is engaged only when a spike is observed, else no information will pass through. This is done to introduce sparse behavior and to retain the energy-saving properties of the LIF neuron. The activation function can be linear or nonlinear, for example, linear activation, Rectified Linear Unit (ReLU), Gaussian Error Linear Unit (GELU), hyperbolic tangent function (tanh), etc. A schematic for the information flow in VSN is shown in Fig. \ref{figure: fc 2}. Note that the amplitude of the spike varies in the proposed VSN as opposed to vanilla SN.

\begin{figure}[ht!]
    \centering
    \includegraphics[width = 0.6\textwidth]{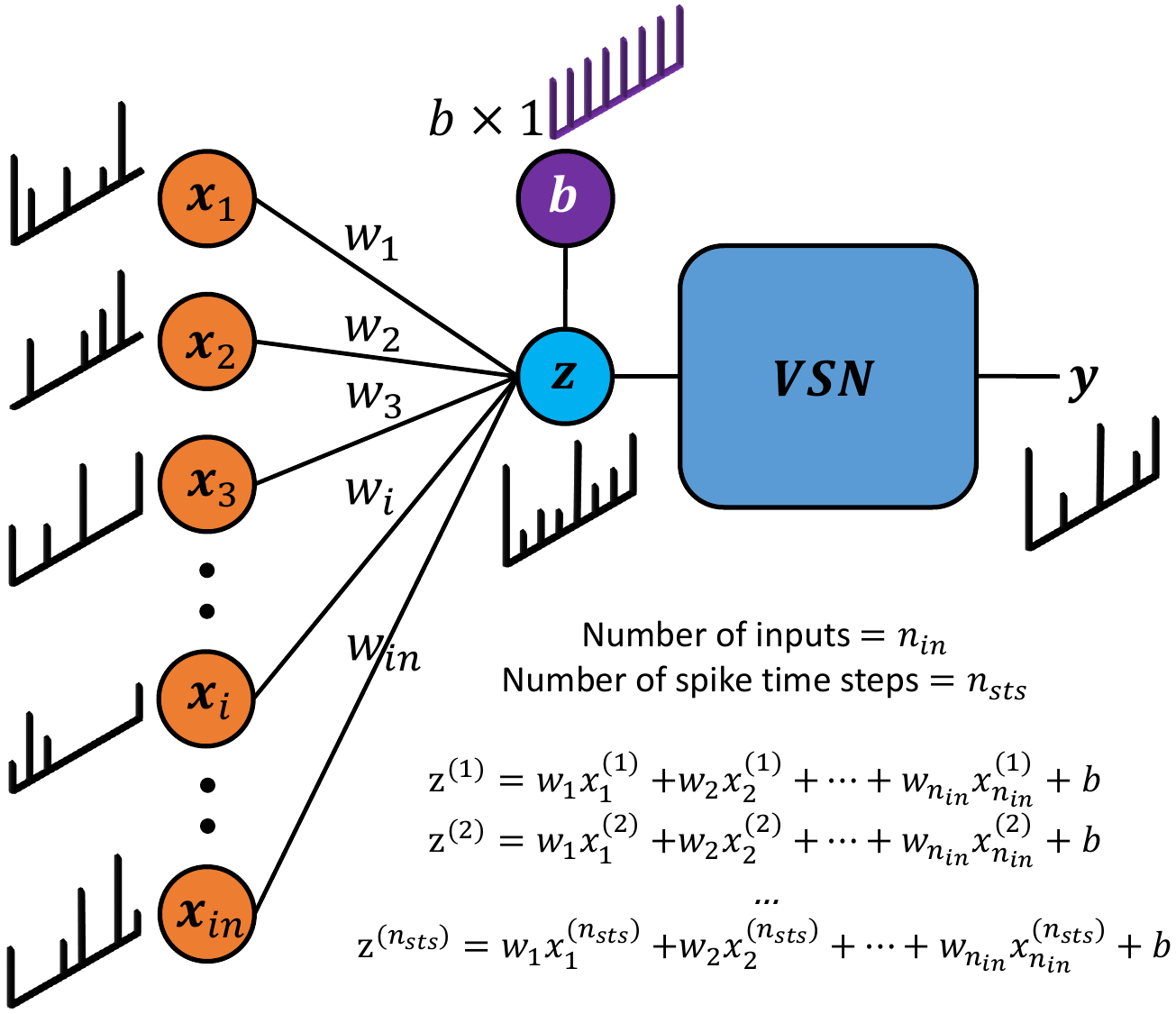}
    \caption{Schematic of the information flow from densely connected linear layer to VSN.}
    \label{fig: dl_VSN}
\end{figure}

Spiking neurons in a neural network are placed in the place of activations. The input $\bm z$ can be obtained from conventional layers like a convolution layer or a densely connected linear layer. The number of spiking neurons in the spiking layer will depend on the dimension of the input data. A schematic for the placement of VSN is shown in Fig. \ref{fig: dl_VSN}.
For the proposed VSN neuron, to compute the derivative of outputs with respect to inputs, i.e., ${d y^{(i)}}/{d z^{(j)}},\,\,\,\,\,j\leq i,\,\,\,\,\, i,j\in\mathbb N$, following equations can be used,
\begin{equation}
\begin{aligned}
\dfrac{d y^{(i)}}{d z^{(j)}} &= \dfrac{\partial y^{(i)}}{\partial \widetilde y^{(i)}}\dfrac{\partial \widetilde y^{(i)}}{\partial M^{(i)}}\prod_{k=j}^{i-1}\left(\dfrac{\partial M^{(k+1)}}{\partial M^{(k)}}+\dfrac{\partial M^{(k+1)}}{\partial \widetilde y^{(k)}}\dfrac{\partial \widetilde y^{(k)}}{\partial M^{(k)}}\right)\dfrac{\partial M^{(j)}}{\partial z^{(j)}}, \,\,\,\,j< i,
\\
\dfrac{d y^{(i)}}{d z^{(j)}} &= \dfrac{\partial y^{(i)}}{\partial \widetilde y^{(i)}}\dfrac{\partial \widetilde y^{(i)}}{\partial M^{(i)}}
\dfrac{\partial M^{(i)}}{\partial z^{(i)}}
+\dfrac{\partial y^{(i)}}{\partial z^{(i)}},\,\,\,\,j=i,
\end{aligned}
\end{equation}
To compute ${\partial \widetilde y^{(j)}}/{\partial M^{(j)}}$ surrogate gradients will be required. This is where surrogate backpropagation \cite{neftci2019surrogate,eshraghian2021training} comes into the picture. Once we have gradients of outputs w.r.t. inputs for spiking neurons, based on the origin of incoming data (ex., densely connected linear layer or convolutional layer), we can compute gradients with respect to weights and bias of the neural network.

\noindent\textbf{Remark: Gradient computations for a densely connected node utilizing VSN as activation}\\ Assume that we have a dense linear layer with $n_{in}$ input nodes before the spiking neuron. Also, there are $n_{sts}$ STS in the incoming spike train, then $\mathbf z$ can be computed as follows,
\begin{equation}
\begin{aligned}
    \mathbf z &= \mathbf w^\mathbf T \mathbf X+b\{\bm 1\}_{1\times n_{sts}},\\
    \mathbf z &= \left[\begin{matrix}z^{(1)} & z^{(2)} & z^{(3)} & ... & z^{(n_{sts})}\end{matrix}\right],\\
    \mathbf w &= \left[\begin{matrix} w_1 & w_2 & w_3 & ... & w_{n_{in}}\end{matrix}\right]^\mathbf T,\\
    \mathbf X &= \left[\begin{matrix} \mathbf x_1 & \mathbf x_2 & ...& \mathbf x_{n_{in}}\end{matrix}\right]^\mathbf T,\\
    \mathbf x_i &= \left[\begin{matrix}
        x_i^{(1)} & x_i^{(2)} & x_i^{(3)} & ... & x_i^{(n_{sts})}\end{matrix}\right],
\end{aligned}
\end{equation}

where $\mathbf X\in \mathbb R^{n_{in}\times n_{sts}}$ is the incoming data for the layer preceding the spiking neuron, $\mathbf w\in \mathbb R^{n_{sts}\times 1}$ is the weight vector, and $b$ is the bias.
The gradients $d\mathbf z/d\mathbf x_i$, $d\mathbf z/d\mathbf w$, $d\mathbf z/d b$ can then be computed as,
\begin{equation}
\begin{aligned}    
    \dfrac{d\mathbf z}{d\mathbf x_i} &= w_i\,\mathbb I_{n_{sts}\times n_{sts}},\\
    \dfrac{d\mathbf z}{d\mathbf w} &= \mathbf X^{\mathbf T},\\
    \dfrac{d\mathbf z}{db} &= \{\mathbf 1\}_{1\times n_{sts}}.
\end{aligned}
\end{equation}\\

\subsection{Energy consumption of spiking neurons:}
The inspiration behind the spiking neurons is sparse communication, as it is believed that the information pipeline \textit{in vivo} conserves energy through the same mechanism. Sparse communications lead to reduced computations, leading to energy consumption. But to answer what level of spiking activity will lead to energy consumption, a detailed analysis is required. Looking at a node in a neural network architecture, we can generalize a few operations that can contribute towards energy consumption. These are (i) retrieving parameters, (ii) synaptic operations, and (iii) activation. Synaptic operations here imply the operations involved in the computation of output given some input. Among these costs, the cost of retrieving parameters and passing activations is largely based on implementation and, as such, is non-trivial to generalize. However, The energy consumption in synaptic operations is largely based on mathematical operations and thus can be generalized well. Discussions in \cite{davidson2021comparison} show that within synaptic operations, four different operations are involved: (i) getting neuron state, (ii)  multiplication, (iii) addition, and (iv) writing neuron state. Based on energy estimation data from post-layout analysis of SpiNNaker2, authors in \cite{davidson2021comparison} take the energy of multiplication and reading neuron state to be 5E. E here is the energy consumed in addition operation and is equal to the energy consumed in the writing neuron state. Through this, it is shown that ANN consumes an energy of 12E$\times N_{\text{mean targets}}$. $N_{\text{mean targets}}$ here represents the target nodes for the current node under consideration. 

Now, extending this analysis to variable spiking neurons, because of variable amplitude spikes, the VSN consumes all the above-stated energies; however, the same is consumed only when a spike is observed. The energy $E_{VSN-Syn}$, consumed in synaptic operations of a VSNN for a particular node, can be computed as,
\begin{equation}
    E_{VSN-Syn} = \text{12E}\times N_{\text{mean targets}}\times N_{\text{avg spikes}},
    \label{eqn: energy}
\end{equation}
where $N_{\text{avg spikes}}$ represents the spikes produced on average in a spike train at a particular node. Comparing this with the energy consumed in AN's synaptic operations, we can conclude that there will be energy savings observed if we have average spikes produced less than one. For this reason, in the following sections, we will report spiking activity to represent energy savings. This is further supported by the fact that the energy consumed in retrieving parameters and passing activations will also depend on spiking activity. This observation is consistent with the literature \cite{davidson2021comparison,dampfhoffer2022snns,lemaire2022analytical} on energy consumption of spiking neurons. The authors would like to note here that although the energy consumed in SNs is less (because of the elimination of multiplication operation) as compared to both VSN and AN, its applicability is also limited, especially in regression tasks. Therefore, it is worthwhile to explore neuron models that introduce sparsity in communication while performing the intended task to satisfaction.
\section{Numerical Illustrations}\label{section: case studies}
To test the proposed VSN, four different examples have been carried out, catering to both classification and regression tasks. SNNTorch \cite{eshraghian2021training} and PyTorch \cite{paszke2019pytorch} are used as deep learning packages. It should be noted that to train the VSNNs and the SNNs in the following examples, surrogate backpropagation \cite{neftci2019surrogate,eshraghian2021training} is used. During backpropagation, the threshold function of the LIF neuron and VSN is idealized using a fast sigmoid \cite{zenke2018superspike,eshraghian2021training} function with a slope parameter equal to 25. The nomenclature for various networks followed while discussing the following results is given in Table \ref{tab:nomenclature}.

\begin{table*}[ht!]
\centering
\caption{Nomenclature for various network architectures.}
\label{tab:nomenclature}
\vspace{1em}\begin{tabular}{p{0.20\textwidth}p{0.65\textwidth}}
    \toprule
    Legend & Detail\\
    \midrule
    ANN & Artificial Neural Network\\
    
    VSNN-\# & VSNN with \# STSs, linear activation in VSN, and no input encoding.\\
    
    VSNN-\#-ReLU & VSNN with \# STSs, ReLU activation in VSN, and no input encoding.\\
    
    SNN-\# & SNN with \# STSs and no input encoding.\\
    
    SNN-\#-RE & SNN with \# STSs and rate encoding as input encoding.\\
    
    SNN-\#-TE & SNN with \# STSs and triangular encoding \cite{kahana2022function,zhang2022sms} as input encoding.
    \\\bottomrule
\end{tabular}
\end{table*}

The average \% spikes ($\tilde S$) reported in the following results are defined for a single layer of SNN or VSNN. For a single spiking layer containing multiple spiking neurons, $\tilde S$ is computed as follows:
\begin{equation}
\tilde S = 100\,\dfrac{\text{total number of spikes produced in the layer over all STSs}}{\text{number of STSs $\times$ number of possible spikes in one STS}}
\end{equation}
Energy consumption of spiking networks is directly proportional to $\tilde S$, as discussed in the previous section. Normalized Mean Square Error (NMSE) values and the accuracy reported in the following examples are computed by comparing the predictions from the trained networks and the ground truth from the particular dataset. All networks used to generate the results in the following examples were trained five times with different random seeds each time, and the results shown report the mean and standard deviation (std.) of the results from the five trials in the form $\underset{\pm std.}{mean}$. 

It should be noted that hyperparameters of neural networks, in different examples, were tuned with reasonable experimentation and exploration. A learning rate of 0.001 was selected for VSNNs and ANN, while a learning rate of 0.0001 was selected for the SNNs. 500 epochs were used to train the classification examples, while 1000 epochs were used to train the two regression examples. A batch size of 200 is taken for classification examples, and 1000 is taken for regression examples. ADAM optimizer was used to train all the networks with a weight decay of $10^{-4}$. Input data is normalized for all four examples, while the outputs are also normalized for the cases with triangular encoding. 
\subsection{Classification}
\subsubsection{MNIST}The first example tackles the classification task by analyzing the MNIST dataset \cite{deng2012mnist}. The architecture selected for various NNs is as follows,
\begin{equation*}
    I(784)\rightarrow DL(200)\rightarrow A1\rightarrow DL(200)\rightarrow A2\rightarrow DL(10) \rightarrow softmax,
\end{equation*}
where $I$ is the input layer, $DL$ is a densely connected linear layer, $A\#$ is the activation used between layers, and $softmax$ \cite{nwankpa2018activation,bridle1989training} is the final output activation. $(\cdot)$ showcases the number of nodes in the layer. In ANN, GELU \cite{hendrycks2016gaussian} is used as the activation function. In VSNNs and SNNs, VSN and LIF neurons replace the activation function of the activation A\#, respectively. The number of spiking neurons in activation  A\# will depend on the input size, and correspondingly, output from these activations will be of the same size as the input. The average \% spikes $\tilde S$ shown in Tables \ref{table: mnist zero}, \ref{table: mnist ablation} and \ref{table: mnist TDS} are for spikes produced in the first activation ($A1$) and the second activation ($A2$). For networks with multiple STSs, the spikes produced after $A2$ in all STSs are collected, and their mean value is forwarded to the last dense layer for further computation.

\begin{table}[ht!]
\caption{MNIST example results produced using various networks with spiking neuron parameters $\beta = 0.90$, and $T = 0.25$. The networks are trained for 50000 Training Samples (TS), and 10000 samples are used for prediction.}
\label{table: mnist zero}
\centering
\vspace{1em}\begin{tabular}{llccccc}
\toprule
\multicolumn{2}{l}{Network} & ANN  & VSNN-1  & VSNN-1-ReLU  & SNN-1  & SNN-10-RE  \\
\midrule
\multicolumn{2}{l}{Accuracy \%} & $\underset{\pm 0.11}{98.03}$  & $\underset{\pm 0.12}{98.05}$ & $\underset{\pm 0.17}{97.92}$ & $\underset{\pm 0.20}{95.67}$ & $\underset{\pm 0.07}{97.89}$
\\[0.9em]
\multicolumn{1}{l}{\multirow{2}{*}{$\tilde S$\%}} & $A1$ & - & $\underset{\pm 0.41}{11.65}$ & $\underset{\pm 0.39}{11.11}$ & $\underset{\pm 0.41}{12.50}$ & $\underset{\pm 0.50}{08.32}$ 
\\[0.9em]
\multicolumn{1}{l}{} & $A2$ & - &  $\underset{\pm 1.16}{23.23}$ &  $\underset{\pm 0.98}{22.13}$ &  $\underset{\pm 1.54}{42.97}$ &  $\underset{\pm 0.45}{15.33}$
\\\bottomrule
\end{tabular}
\end{table}
Table \ref{table: mnist zero} shows the prediction accuracy achieved and the average \% spikes produced for MNIST example using various networks. It can be seen that the performance of VSNNs using VSNs paired with either linear or ReLU activation is at par with the SNNs and ANNs. The average spikes produced in VSNNs are less than those produced in SNN-1. Also, while spikes produced per STS are less in the case of the SNN-10-RE network, the total number of STSs required is more, with no gain in accuracy compared to VSNNs.

\begin{table}[ht!]
\caption{Various case studies for MNIST example. The network under consideration is VSNN-1, with leakage parameter $\beta = 0.90$. $\#$ marked case only converged for two out of five trials.}
\begin{subtable}[ht!]{0.475\textwidth}
\caption{Ablation study results. The network is trained using 50000  Training Samples (TS), and 10000 samples are used for prediction.}.
\label{table: mnist ablation}
\centering
\begin{tabular}{llcccc}
\toprule
\multicolumn{2}{l}{T} & 0.025  & 0.10  & 0.25 & 0.30$^\#$
\\\midrule
\multicolumn{2}{l}{Accuracy \%} & $\underset{\pm 0.05}{98.06}$  & $\underset{\pm 0.06}{97.99}$ & $\underset{\pm 0.12}{98.05}$ & $\underset{\pm 0.36}{97.75}$
\\\midrule
\multicolumn{1}{l}{\multirow{2}{*}{$\tilde S$\%}} & $A1$ & $\underset{\pm 0.43}{15.95}$ & $\underset{\pm 0.44}{14.86}$ & $\underset{\pm 0.41}{11.65}$ & $\underset{\pm 2.26}{08.95}$ 
\\[0.9em]
\multicolumn{1}{l}{} & $A2$ &  $\underset{\pm 1.85}{20.86}$ &  $\underset{\pm 0.93}{20.73}$ &  $\underset{\pm 1.16}{23.23}$ &  $\underset{\pm 0.10}{24.61}$
\\\bottomrule
\end{tabular}
\end{subtable}
\hfill
\begin{subtable}[ht!]{0.475\textwidth}
\caption{Training data study results. Threshold parameter, $T = 0.25$. 10000 samples
are used for prediction.\\}
\label{table: mnist TDS}
\centering
\begin{tabular}{llcccc}
\toprule
\multicolumn{2}{l}{TS($10^3$)} & 5  & 12.5  & 25 & 50
\\\midrule
\multicolumn{2}{l}{Accuracy \%} & $\underset{\pm 0.23}{94.73}$  & $\underset{\pm 0.69}{96.19}$ & $\underset{\pm 0.10}{97.63}$ & $\underset{\pm 0.12}{98.05}$
\\\midrule
\multicolumn{1}{l}{\multirow{2}{*}{$\tilde S$\%}} & $A1$ & $\underset{\pm 3.17}{25.39}$ & $\underset{\pm 1.36}{14.99}$ & $\underset{\pm 0.95}{13.10}$ & $\underset{\pm 0.41}{11.65}$ 
\\[0.9em]
\multicolumn{1}{l}{} & $A2$ &  $\underset{\pm 3.74}{50.01}$ &  $\underset{\pm 3.97}{47.26}$ &  $\underset{\pm 1.89}{41.71}$ &  $\underset{\pm 0.16}{23.23}$
\\\bottomrule
\end{tabular}
\end{subtable}
\end{table}

To test the efficacy of VSNNs, the effect of changing the neuron parameters on accuracy were also tested. Table \ref{table: mnist ablation} shows the accuracy achieved for various values of threshold parameters in the VSNN-1 network. It can be seen that the threshold dictates the spiking activity for the VSNs, and if a very high value of the threshold is selected, the network may fail to map the training data entirely. A threshold value of 0.50 was also tested, but the network failed to converge since there was insufficient neuron activity. It should be noted that all the spiking neurons, i.e., the 200 in the first activation and the 200 in the second activation, were assigned the same values of $\beta$ and $T$ while running any specific case. Because we are only taking a single STS for training the VSNN network, leakage parameter $\beta$ has no effect on the final results.

Table \ref{table: mnist TDS} shows the results of training data studies carried out for the MNIST example. As can be seen, as the Training Samples (TS) increase, the accuracy achieved also increases, which is an expected behavior. Interestingly, the spiking activity reduces (even in the prediction stage with an increase in training data). The possible explanation for this is that the network can optimize its trainable parameters better with the increase in training data. 

\subsubsection{Fashion-MNIST} The second classification example explores the performance of VSN for Fashion-MNIST \cite{xiao2017online} dataset. The network architecture is as follows:
\begin{multline*}
I(9)\rightarrow CL(10,3)\rightarrow A1\rightarrow CL(30,3)\rightarrow A2\rightarrow MP \rightarrow\\
CL(10,3) \rightarrow A3\rightarrow CL(10,3)\rightarrow A4\rightarrow F \rightarrow \\ DL(320)\rightarrow A5\rightarrow DL(10)\rightarrow softmax,
\end{multline*}
where $CL(\#1,\#2)$ denotes a convolution layer with $\#1$ output channels and $\#2$ as the square kernel size. $F$ denotes a flattening layer. ReLU is used as the activation for the ANN. For spiking networks, a procedure similar to the previous example is followed. In direct input cases with multiple STS, input nodes at each spike time receive the same input.

\begin{table}[ht!]
\caption{Fashion-MNIST example results produced using various networks with the spiking neuron parameters, $\beta = 0.90$, and $T = 0.05$. The networks are trained using 50000 TS, and 10000 samples are used for prediction. SNN-1 failed to converge for the Fashion-MNIST dataset.}
\label{table: Fashion MNIST}
\centering
\vspace{1em}\begin{tabular}{llcccc}
\toprule
\multicolumn{2}{l}{Network} & ANN  & VSNN-1  & VSNN-5 & SNN-5
\\\midrule
\multicolumn{2}{l}{Accuracy \%} & $\underset{\pm 0.19}{89.49}$  & $\underset{\pm 0.30}{89.34}$ & $\underset{\pm 0.33}{89.98}$ & $\underset{\pm 0.37}{83.25}$
\\\midrule
\multicolumn{1}{l}{\multirow{5}{*}{$\tilde S$\%}} & $A1$ & \multicolumn{1}{l}{\multirow{5}{*}{-}} &  $\underset{\pm 8.10}{59.99}$ &$\underset{\pm 3.72}{38.54}$ & $\underset{\pm 0.05}{1.72}$ 
\\[0.9em]
\multicolumn{1}{l}{} & $A2$ & & $\underset{\pm 4.54}{32.01}$ &  $\underset{\pm 3.57}{21.55}$ & $\underset{\pm 0.68}{2.89}$
\\[0.9em]
\multicolumn{1}{l}{} & $A3$ & & $\underset{\pm 11.07}{56.64}$ &  $\underset{\pm 6.69}{36.09}$ &  $\underset{\pm 1.08}{14.97}$
\\[0.9em]
\multicolumn{1}{l}{} & $A4$ & & $\underset{\pm 1.71}{23.33}$ &  $\underset{\pm 1.56}{14.77}$ &  $\underset{\pm 0.93}{15.55}$
\\[0.9em]
\multicolumn{1}{l}{} & $A5$ & & $\underset{\pm 0.34}{7.72}$ &  $\underset{\pm 0.34}{4.92}$ &  $\underset{\pm 0.26}{24.24}$
\\\bottomrule
\end{tabular}
\end{table}
Table \ref{table: Fashion MNIST} shows the accuracy achieved in various networks for the Fashion-MNIST example. The results produced using the VSNN-1 network are at par with the ANN and are better than the SNNs despite only considering a single STS. Now, although the spikes produced per STS are more in VSNN-1 as compared to SNN-5, it should be noted that VSNN-1 converged in a single STS, and also the accuracy achieved in VSNN is more than that observed in SNN-5 network. As for the VSNN-5 network, even though we observe a slight increase in accuracy, the total spikes produced across all STSs will result in more energy consumption, thus defeating the intended purpose.
\subsection{Regression}
For regression, two examples are shown, testing the efficacy of the proposed VSNs. The datasets are selected from Penn Machine Learning Benchmarks \cite{Olson2017PMLB} collection. The first dataset selected is the \textit{feynman\_I\_6\_2a} dataset \cite{EpistasisLab_1feat} with single input and output feature. The dataset is from here on referred to as \textit{1-feature-dataset}. The second dataset selected is the \textit{feynman\_I\_9\_18} dataset \cite{EpistasisLab_9feat} with nine input features and a single output feature.
The dataset is from here on referred to as the \textit{9-feature-dataset}.

\subsubsection{1-feature-dataset}
The network architecture for the 1-feature-dataset is as follows,
\begin{multline*}
    I(1)\rightarrow DL(100)\rightarrow A1\rightarrow DL(250)\rightarrow\\ A2\rightarrow DL(500)\rightarrow A3\rightarrow DL(250)\rightarrow A4\rightarrow \\DL(100)\rightarrow A5\rightarrow DL(1).
\end{multline*}
GELU is used as the activation for the ANN. For VSNNs and SNNs, in the case of multiple STS, the spike train generated after the last activation is stored and decoded before forwarding to the last dense layer for final output. For direct encoding, the decoder is replaced by a mean function, which reduces the spike train by taking the mean value.
\begin{table}[ht!]
\caption{NMSE values obtained from various networks for 1-feature-dataset example. Spiking neuron parameters $T = 0.01$ and $\beta = 0.90$. 4000 TS are used for training, and 96000 samples are used for prediction.}
\label{table: regression-1 zero}
\centering
\vspace{1em}\begin{tabular}{lcccc}
\toprule
\multicolumn{1}{l}{Network} & ANN  & VSNN-1  & SNN-50-TE  & SNN-100-TE  \\\midrule
\multicolumn{1}{l}{NMSE ($\times10^{-4}$)} & $\underset{\pm0.8}{8.8}$ & $\underset{\pm8.3}{12.4}$ & $\underset{\pm114.0}{145.0}$ & $\underset{\pm168.0}{136.0}$
\\\bottomrule
\end{tabular}
\end{table}
Table \ref{table: regression-1 zero} shows the NMSE values for the 1-feature-dataset example. As can be seen, the results produced using the VSNN-1 network are at par with the ANN and are far better than SNNs despite only considering a single STS.
For the triangular encoding case, input nodes received inputs according to the encoded spike train. It was observed that for VSNN-1 network, in the hidden layers, after $A1$, $A2$, $A3$, $A4$ and $A5$, on average, 22.73\%, 12.95\%, 6.87\%, 13.32\%, and 33.72\% spikes were produced. This can result in energy saving, compared to conventional ANNs, while preserving prediction accuracy. 

\subsubsection{9-feature-dataset} The network architecture for the 9-feature-dataset is as follows,
\begin{multline*}
    I(9)\rightarrow DL(100)\rightarrow A1\rightarrow DL(150)\rightarrow\\ A2\rightarrow DL(250)\rightarrow A3\rightarrow DL(150)\rightarrow\\ A4\rightarrow DL(100)\rightarrow A5\rightarrow DL(1),
\end{multline*}
GELU is used as the activation for the ANN. For spiking networks with multiple STSs or direct encoding, a procedure similar to the previous example is followed.

\begin{table}[ht!]
\caption{Regression 9-feature-dataset example results for various networks. Spiking neuron parameters, $\beta = 0.90$, and $T = 0.01$. NMSE values are reported for predictions carried out on 95000 samples (5000 TS were used for training).}
\label{table: regression-9 zero subtract}
\centering
\vspace{1em}\begin{tabular}{lccccc}
\toprule
\multicolumn{1}{l}{Network} & ANN  & VSNN-1 & VSNN-1-ReLU  & SNN-50  & SNN-100-TE  \\\midrule
\multicolumn{1}{l}{NMSE ($\times10^{-4}$)} & $\underset{\pm 1.4}{12.2}$ & $\underset{\pm 0.4}{4.8}$ &
$\underset{\pm 0.9}{5.3}$ & $\underset{\pm 15.1}{83.1}$ & $\underset{\pm 48.0}{89.2}$
\\\bottomrule
\end{tabular}
\end{table}
Table \ref{table: regression-9 zero subtract} showcases the result for networks trained for the 9-feature-dataset example. It can be seen that the performance of variable spiking neural networks using VSN paired with either linear or ReLU activation is superior to both the SNNs and ANNs. While the ANN was able to generate reasonably good results, SNNs did not train well for both encoded and direct input cases.  

\begin{table}[ht!]
\caption{The values reported are the average \% spikes produced after various activations of the 9-feature-dataset example. The VSNN networks are trained using 5000 TS, and predictions are made on 95000 samples. $T= 0.01$.}
\label{table: regression-9 zero spikes}
\centering
\vspace{1em}\begin{tabular}{lccccc}
\toprule
\multicolumn{1}{l}{Layers} & $A1$  & $A2$  & $A3$  & A4  & A5  \\\midrule
\multicolumn{1}{l}{VSNN-1 $\tilde S$\%}& $\underset{\pm2.59}{40.32}$ 
& $\underset{\pm1.91}{14.48}$ & $\underset{\pm1.71}{11.68}$ & $\underset{\pm4.98}{23.10}$ & $\underset{\pm3.48}{39.09}$ \\[0.9em]
\multicolumn{1}{l}{VSNN-1-ReLU $\tilde S$\%} & $\underset{\pm4.09}{41.43}$ 
& $\underset{\pm2.15}{14.16}$ & $\underset{\pm2.05}{9.49}$ & $\underset{\pm4.35}{23.73}$ & $\underset{\pm2.27}{41.14}$
\\\bottomrule
\end{tabular}
\end{table}
Table \ref{table: regression-9 zero spikes} shows the spikes produced in the trained VSNNs while predicting for the 9-feature-dataset example. Similar to the previous example, the spiking neurons in the activation layer only activate sparingly. The comparison with SNNs is not shown since they failed to converge to ground truth. 

\begin{table}[ht!]
\caption{Ablation study NMSE values reported for 9-feature-dataset example. VSNN-1-ReLU network is trained for the study using 5000 TS. 95000 samples are used for prediction.}
\label{table: regression-9 ablation}
\centering
\vspace{1em}\begin{tabular}{lccccc}
\toprule
\multicolumn{1}{l}{Threshold} & 0.005  & 0.010  & 0.020  & 0.025 & 0.050\\\midrule
\multicolumn{1}{l}{NMSE ($\times10^{-4}$)} & $\underset{\pm3.8}{5.5}$ 
& $\underset{\pm0.9}{5.3}$ & $\underset{\pm1.0}{8.3}$ & $\underset{\pm2.5}{13.3}$ & $\underset{\pm83}{112}$
\\\bottomrule
\end{tabular}
\end{table}
\begin{figure}[ht!]
\begin{subfigure}{0.49\textwidth}
    \centering
    \includegraphics[width = \textwidth]{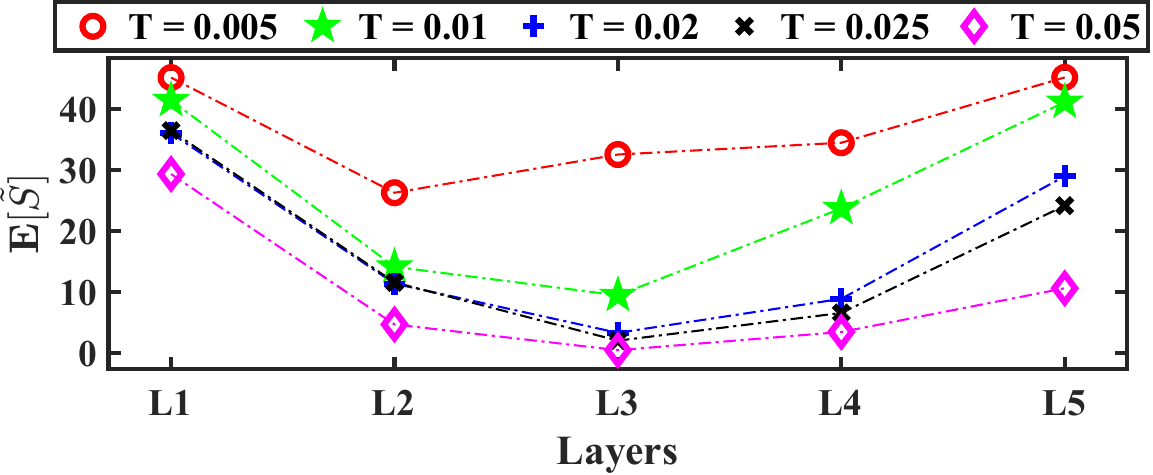}
    \caption{\centering Average \% spikes produced in ablation studies for 9-feature-dataset.}
    \label{fig:9FD_AS}
\end{subfigure}
\begin{subfigure}{0.49\textwidth}
    \centering
    \includegraphics[width = \textwidth]{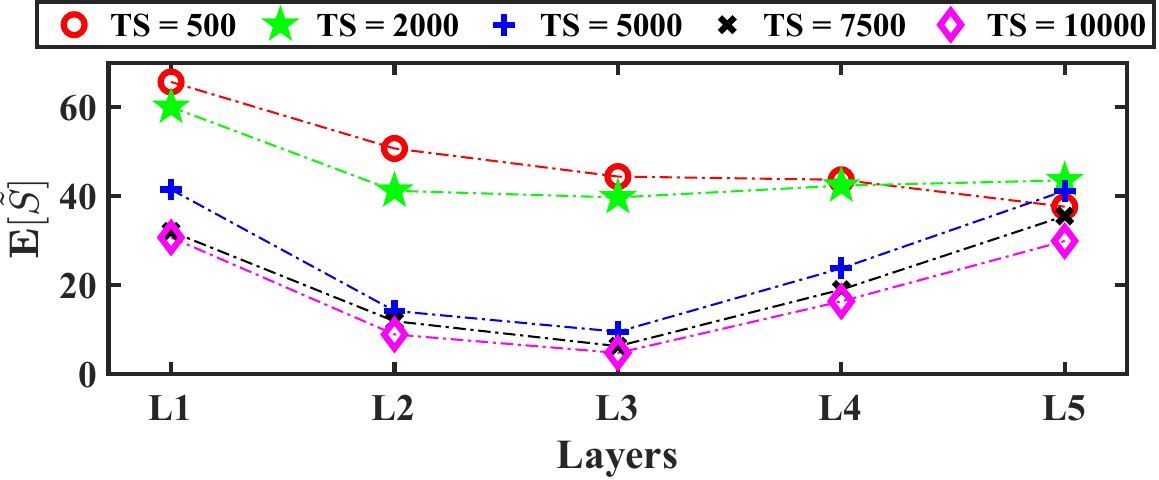}
    \caption{\centering Average \% spikes produced in training data studies for 9-feature-dataset.}
    \label{fig:9FD_TDS}
\end{subfigure}
\caption{Average \% spikes produced in various studies carried out for 9-feature-dataset.}
\end{figure}
The accuracy of the VSNN-ReLU network for the 9-feature-dataset example was also tested for different threshold values. Since only a single time step is considered, the leakage parameter will not affect the accuracy. Table \ref{table: regression-9 ablation} shows the NMSE values corresponding to different thresholds. It can be seen that as we increase the threshold, the accuracy drops, which can be attributed to reduced spiking activity of the spiking neurons. However, up to a certain value of threshold (which will differ for different datasets), 0.02 for the current example, the accuracy is not affected by a lot, whereas the spiking activity reduces considerably (refer Fig. \ref{fig:9FD_AS}). Hence, for different datasets, it is worthwhile to test for an appropriate threshold value, which results in maximum savings while generating fairly accurate results.

\begin{table}[ht!]
\caption{NMSE values observed during training data studies for the 9-feature-dataset example. VSNN-1-ReLU network is trained for the study. $T= 0.01$. 90000 samples were used for prediction.}
\label{table: regression-9 TDS}
\centering
\vspace{1em}\begin{tabular}{lccccc}
\toprule
\multicolumn{1}{l}{Training Samples} & 500  & 2000  & 5000  & 7500 & 10000\\\midrule
\multicolumn{1}{l}{NMSE ($\times10^{-4}$)} & $\underset{\pm1.0}{14.2}$ 
& $\underset{\pm0.4}{7.9}$ & $\underset{\pm0.9}{5.3}$ & $\underset{\pm1.3}{6.3}$ & $\underset{\pm0.6}{5.4}$
\\\bottomrule
\end{tabular}
\end{table}
Table \ref{table: regression-9 TDS} shows the NMSE values observed during training data studies for the 9-feature-dataset example. VSNN-1-ReLU network is trained for the study. Similar to the MNIST example, as the training data increases, the accuracy increases, and the spiking activity reduces (ref Fig. \ref{fig:9FD_TDS}). In the current example for ten thousand training samples, we observe the least spiking activity while the NMSE observed is less than that observed for ANN.

\begin{figure}[ht!]
    \centering
    \includegraphics[width = 0.75\linewidth]{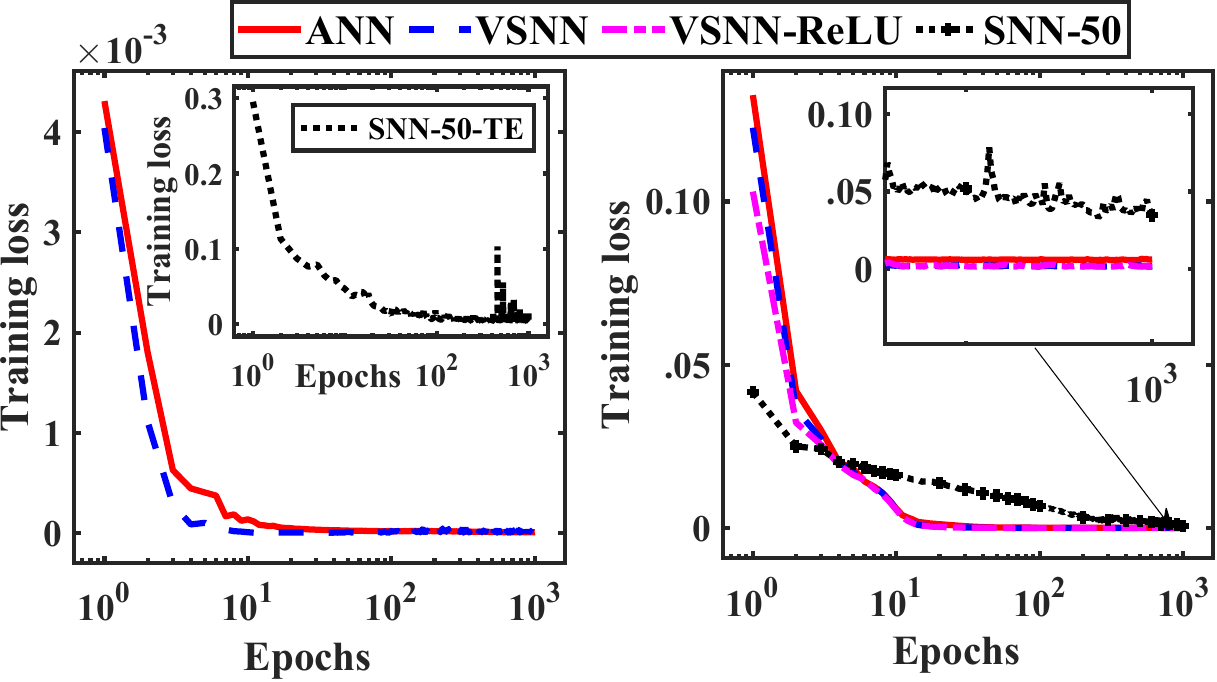}
    \caption{Epoch vs training loss for 1-feature-dataset (left) and 9-feature-dataset (right).}
    \label{fig:epoch_training_loss}
\end{figure}

Fig. \ref{fig:epoch_training_loss} shows the evolution of training loss with epochs for regression examples. The results shown are for a single run of networks. As can be seen, the VSN's behavior is similar to that observed in ANN, while SNN's behavior is more jittery and does not follow the same pattern. SNN-50-TE results are shown separately as output data was normalized; hence, the mean square error's magnitude will be different.
A comparison of energy consumption of various examples is shown in Appendix \ref{sec: energy}. Furthermore, in Appendix \ref{sec:Additional}, we discuss some additional numerical illustrations. Appendix \ref{sec: Add pI} shows the effects of multiple STS in the VSNN-ReLU network, trained for 9-feature-dataset, and Appendix \ref{sec: Add pII} shows an additional regression example, which requires multiple STS to converge.  
\section{Conclusion}\label{section: conclusion}
In this paper, we proposed a novel nonlinear spiking neuron that combines the properties of ANs and SNs. It gains the property of continuous activation from AN and gains sparsity in communication from SN. Furthermore, it reduces the disadvantages of both the parent neuron models, as is observed from the results. The gain in sparsity introduced by VSN, which was a weak point for AN, well compensates for the slight dip in the prediction accuracy, which was a weak point for SNs. However, the authors here note that in the ranks of energy-efficient neurons, VSN is placed between SN and AN and is not the most energy-efficient neuron.

From the results produced, it can be seen that the VSN is well-suited for regression tasks; however, it also performs well for classification tasks. The various examples shown in the previous section show that the accuracy of VSNNs is at par with the accuracy achieved using ANNs and is even better in the 9-feature-dataset example. The advantages observed in neural networks utilizing VSNs include,
\begin{enumerate}
    \item The ability to produce accurate results with a few STSs. In the examples shown, the VSN was able to converge to ground truth in a single time step.
    \item The ability to produce good results with non-encoded raw input data.
    \item The ability to tackle regression tasks for both multi-input and single-input datasets while promoting sparsity in communication.
\end{enumerate}
The proposed VSNs' ability to spike intermittently can result in huge power savings, especially for cases where neuron parameters like $\beta$ and $T$ are tuned properly.

Having discussed the advantages of the proposed variable spiking neurons, the authors note that there is a huge scope for further research in this area. VSN's performance on neuromorphic hardware is yet to be explored, along with its performance on complex neural network architectures like neural operators. Research can also be carried out to explore the possible ways for tuning the VSN's parameters such that the spiking activity is minimized and the accuracy achieved is maximized. Also, neuroscience-based deep learning as a whole can benefit greatly from a training algorithm designed specifically for training neural networks utilizing spiking neurons. 

\section*{Acknowledgment}
SG acknowledges the financial support received from the Ministry of Education, India, in the form of the Prime Minister's Research Fellows (PMRF) scholarship. SC acknowledges the financial support received from the Science and Engineering Research Board (SERB) via grant no. SRG/2021/000467 and from Ministry of Port and Shipping via letter no. ST-14011/74/MT (356529).


\appendix
\section{Energy Estimates}\label{sec: energy}
In this section, we will show the energy consumption in the synaptic operations of VSNNs. An energy metric $E_r$ is introduced, which is the ratio of energy consumed in synaptic operation based on observed spiking activity to the energy consumed assuming 100\% spiking activity. To compute total energy in synaptic operations, given a certain spiking activity, Eq. \eqref{eqn: energy} is used. In this, we need to define the number of mean targets for various conditions. For a layer $l-1$, densely connected to layer $l$ with $N$ nodes, the number of mean targets for each node of layer $l-1$ will be $N$. Similarly, for a convolution layer with an input image ensemble having $C$ channels and individual image size of $L\times H$, a kernel size of $L_k\times H_k$, the number of mean targets $N_m$ for each element of the input ensemble can be computed as follows,
\begin{equation}
    N_m = L_kH_k\dfrac{LH}{L_oH_o},
\end{equation}
where the output image size is denoted as $L_o\times H_o$. The above equation holds for a single stride in both directions and zero padding. The $N_m$ is defined for one output channel; the same will be multiplied by the number of output channels for a multi-channel output case. Table \ref{tab:my_label} shows the $E_r$ values observed in various examples. As can be observed, VSNN, because of sparse communication, conserves energy in comparison to the ANN. The MNIST\textsuperscript{*} case shown has an extra VSN layer after the input, before the first layer with 200 nodes.
\begin{table}[ht!]
    \centering
    \caption{Ratio of energy consumed in synaptic operation based on observed spiking activity to the energy consumed assuming 100\% spiking activity different examples.}  
    \vspace{1em}\begin{tabular}{llc}
    \toprule
    Example & Network & $E_r$ \\\midrule
    \multirow{2}{*}{MNIST} & VSNN-1 & 0.81\\
    & VSNN-1-ReLU & 0.81\\
    \multirow{2}{*}{MNIST\textsuperscript*} & VSNN-1 & 0.15\\
    & VSNN-1-ReLU & 0.15\\
    Fashion-MNIST & VSNN-1 & 0.63\\
    1-feature-dataset & VSNN-1 & 0.11\\
    \multirow{2}{*}{9-feature-dataset} & VSNN-1 & 0.19\\
    & VSNN-1-ReLU & 0.18
    \\\bottomrule
    \end{tabular}
    \label{tab:my_label}
\end{table}
\section{Additional Numerical Illustrations}\label{sec:Additional}
Additional insights related to examples discussed in section 4 of the manuscript are discussed in this section. Furthermore, an additional example is discussed, showcasing the effect of multiple STS on the performance of VSN for regression tasks.
\subsection{9-feature-dataset}\label{sec: Add pI}
\begin{table}[ht!]
\caption{Effect of multiple STS on the VSNN-ReLU network, trained for the 9-feature-dataset. The leakage parameter, $\beta = 0.9$ and the threshold parameter, $T = 0.01$. The network is trained for 5000 training samples and 95000 samples are used for prediction.\\}
\label{table: 9fd msts}
\centering
\vspace{1em}\begin{tabular}{llcccc}
\toprule
\multicolumn{2}{l}{STS} & 1 & 5 & 10 & 20
\\\midrule
\multicolumn{2}{l}{NMSE ($\times10^{-4}$)} & $\underset{\pm 0.9}{5.3}$  & $\underset{\pm 1.9}{6.0}$ & $\underset{\pm 0.8}{5.4}$ & $\underset{\pm 0.4}{5.4}$
\\\midrule
\multicolumn{1}{l}{\multirow{5}{*}{$\tilde S$\%}} & $A1$ & $\underset{\pm 4.09}{41.43}$ &$\underset{\pm 1.13}{28.56}$ & $\underset{\pm 1.50}{23.79}$ & $\underset{\pm 1.02}{24.25}$ 
\\[0.9em]
\multicolumn{1}{l}{} & $A2$ & $\underset{\pm 2.15}{14.16}$ &  $\underset{\pm 1.43}{19.01}$ & $\underset{\pm 2.14}{19.82}$ &  $\underset{\pm 1.88}{19.93}$
\\[0.9em]
\multicolumn{1}{l}{} & $A3$ & $\underset{\pm 2.05}{9.49}$ &  $\underset{\pm 0.91}{23.46}$ & $\underset{\pm 1.44}{20.66}$ &  $\underset{\pm 0.72}{21.90}$
\\[0.9em]
\multicolumn{1}{l}{} & $A4$ & $\underset{\pm 4.35}{23.73}$ &  $\underset{\pm 1.51}{21.87}$ & $\underset{\pm 1.04}{18.66}$ &  $\underset{\pm 0.85}{19.48}$
\\[0.9em]
\multicolumn{1}{l}{} & $A5$ & $\underset{\pm 2.27}{41.14}$ &  $\underset{\pm 1.79}{25.70}$ & $\underset{\pm 0.87}{21.93}$ & $\underset{\pm 0.75}{21.85}$
\\\bottomrule
\end{tabular}
\end{table}
A study showcasing the effect of multiple spike time steps was carried out for the 9-feature-dataset, and the results produced are shown in Table \ref{table: 9fd msts}. As can be seen that the accuracy achieved using the VSNN-ReLU network converged in a single STS, thus eliminating the need to consider multiple STS. A similar trend is observed in all four examples discussed in section 4 of the manuscript.
\subsection{Regression example III}\label{sec: Add pII}
An additional example was carried out to test the performance of the proposed VSN for regression tasks. The data for this example was generated by self and the mapping was carried out from $x$ to $10^{-5}(5x^2+200x)$. The network architecture for the current example is as follows,
\begin{equation*}
    I(1)\rightarrow DL(100)\rightarrow A1\rightarrow DL(200)\rightarrow A2\rightarrow DL(200)\rightarrow A3\rightarrow DL(100)\rightarrow A4\rightarrow DL(1),
\end{equation*}
where $I$ is the input layer, $DL$ is a densely connected linear layer, and $A\#$ is the activation used between layers. $(\cdot)$ showcases the number of nodes in the layer. In ANN, GELU is used as the activation function. In VSNNs and SNNs, VSN and LIF neurons replace the activation function of the activation A\#, respectively. The number of spiking neurons in activation  A\# will depend on the input size, and correspondingly, output from these activations will be of the same size as the input. A learning rate of 0.001 was selected for ANN and VSNN-$\#$ networks, while a learning rate of 0.0001 was selected for the SNN. In direct input cases with multiple STS, input nodes at each spike time receive the same input, while in the triangular encoding case, the input node at each STS receives the relevant input as per the encoded data. A batch size of 200 was selected for the example. Input data was normalized for all the networks. 
\begin{table}[ht!]
\caption{NMSE values observed using various networks for example mapping $x$ to $10^{-5}(5x^2+200x)$. The leakage parameter, $\beta = 0.9$, and the threshold parameter, $T = 0.15$. Two hundred samples were used for training, and 10000 samples were used for prediction.\\}
\label{table: self nmse}
\centering
\vspace{2em}\vspace{1em}\begin{tabular}{llccccc}
\toprule
\multicolumn{2}{l}{Network} & ANN & VSNN-1 & VSNN-2 & VSNN-5 & SNN-50-TE
\\\midrule
\multicolumn{2}{l}{NMSE ($\times10^{-4}$)} & $\underset{\pm 0.1}{0.9}$  & $\underset{\pm 26.4}{24.6}$ & $\underset{\pm 1.4}{5.2}$ & $\underset{\pm 1.0}{3.4}$ & $\underset{\pm 23.5}{47.0}$
\\\bottomrule
\end{tabular}
\end{table}
Table \ref{table: self nmse} shows the Normalized Mean Square Error (NMSE) observed using various networks, trained for the current example. It can be observed that the NMSE values obtained using VSNNs decrease with an increase in STS and approach those observed using ANN. SNN, however, fails to converge despite taking fifty STSs. Authors here note that a different combination of $\beta$ and $T$ may further improve the results. Also, the results may benefit from a training algorithm that is tailored for spiking networks.
\end{document}